\documentclass[10pt, a4paper]{article}
\usepackage{lrec2016}
\usepackage{multibib}
\newcites{languageresource}{Language Resources}
\usepackage{graphicx}
\usepackage{alltt}
\usepackage{verbatim}

\usepackage{epstopdf}
\usepackage[latin1]{inputenc}

\title{PersonaBank: A Corpus of Personal Narratives \\and Their Story Intention Graphs}

\name{Stephanie M. Lukin, Kevin Bowden, Casey Barackman, and
  Marilyn A. Walker}

\address{University of California Santa Cruz \\
Natural Language and Dialogue Systems Lab \\ 
Computer Science Department \\ 
slukin@ucsc.edu, mawalker@ucsc.edu\\} 

\abstract{
We present a new corpus, PersonaBank, consisting of 108 personal
stories from weblogs that have been annotated with their {\sc story
intention graphs}, a deep representation of the fabula of a story.
We describe the topics of the stories and the basis of the
{\sc story intention graph} representation, as well as the process
of annotating the stories to produce the {\sc story intention graph}s and the challenges of adapting the tool to this new personal narrative domain
We also discuss how the corpus can be used in applications that
retell the story using different styles of tellings, co-tellings, or as a content planner.
\\ 
\newline 
 }

\begin{document}

\maketitleabstract

\section{Introduction}

Hundreds of thousands of personal narratives are published on the web
each month in weblogs. These personal narratives provide direct
insight into the daily lives of people, the activities they engage in
and their beliefs, goals and plans. We have developed a new corpus,
PersonaBank, that consists of 108 personal narratives annotated with a
deep representation of narrative called a {\sc story intention graph}
or {\sc sig}. The story topics include stories about romance, travel,
sports, holidays, watching wildlife, and weather. The stories have
also been annotated for overall positive and negative tone. The {\sc
  sig} representation provides a propositional representation of the
story timeline, the goals and motivations of the story characters, and
the affective impacts of story events on characters.

Our approach builds on the
corpus and tools associated with the DramaBank language resource, a
collection of Aesop's Fables and other classic stories that
utilize the {\sc sig} representation
\cite{elson2010automatic,elson2012dramabank,elson2012detecting}.  This work is the
first to apply this formalism to informal personal narratives, such as
the story about {\it The Startled Squirrel} shown in
Figure~\ref{squirrel-blog-story}.

\begin{figure}[htb]
\vspace{-.05in}
\begin{small}
\begin{tabular}{|p{3.0in}|}
\hline 
This is one of those times I wish I had a digital camera. We keep a
large stainless steel bowl of water outside on the back deck for
Benjamin to drink out of when he's playing outside. His bowl has
become a very popular site. Throughout the day, many birds drink out
of it and bathe in it. The birds literally line up on the railing
and wait their turn. Squirrels also come to drink out of it. The
craziest squirrel just came by- he was literally jumping in fright at
what I believe was his own reflection in the bowl. He was startled so
much at one point that he leap in the air and fell off the deck. But
not quite, I saw his one little paw hanging on! After a moment or
two his paw slipped and he tumbled down a few feet. But oh, if you
could have seen the look on his startled face and how he jumped back
each time he caught his reflection in the bowl! \\
\hline
\end{tabular}
\caption{{\it Startled Squirrel} personal narrative \label{squirrel-blog-story}}
\vspace{-.1in}
\end{small}
\end{figure}

The {\sc sig} is an abstract model of
narrative structure that was designed to be able to represent any
story in terms of its characters, and their actions, intentions and
affectual motivations. Part of a {\sc sig} is shown in Figure~\ref{squirrel-sig} and is described in more detail in Section~\ref{sig-sec} 
We believe {\sc sig}s provide
a useful basis for theoretical analyses of narrative structure
and for applications related to language processing
or storytelling. There are several 
practical advantages to using the
{\sc sig} representation for our corpus:

\begin{figure*}[htb]
\begin{center}
\centering
\includegraphics[width=4.5in]{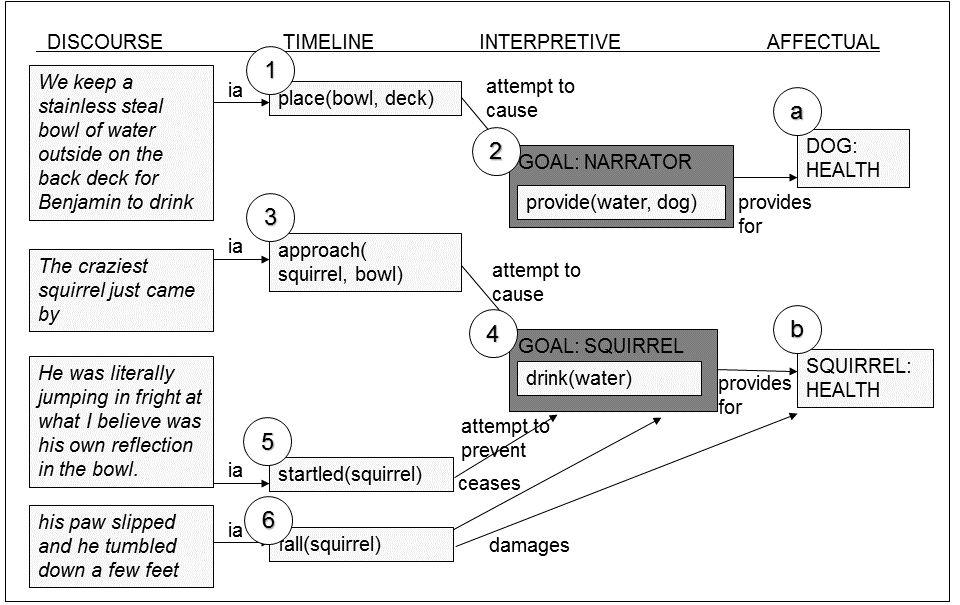}
\vspace{-0.1in}
\caption{\label{squirrel-sig} Part of the {\sc Story Intention Graph} for the {\it Startled Squirrel}}
\vspace{.1in}
\includegraphics[width=4.5in]{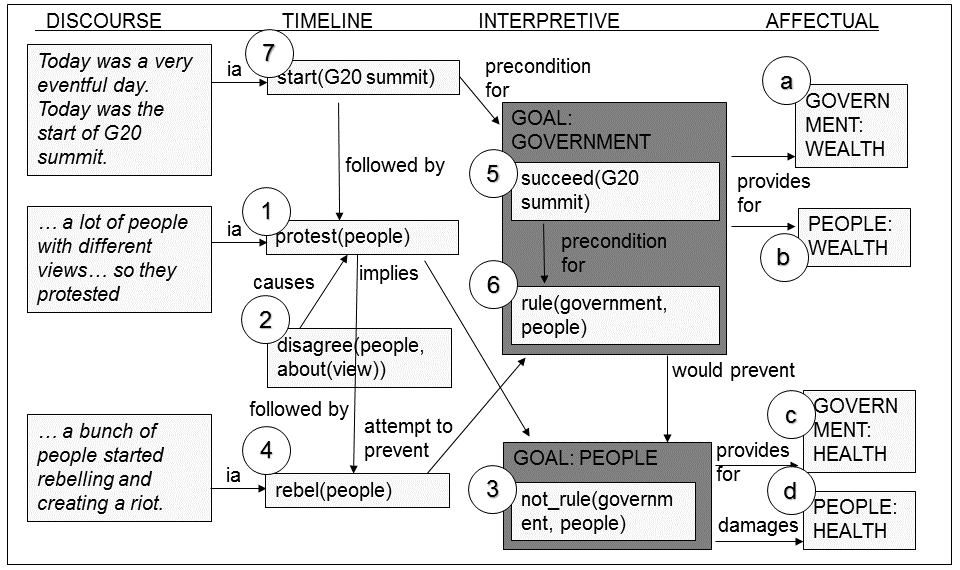}
\vspace{-.1in}
\caption{\label{protest-sig} Part of the {\sc Story Intention Graph} for the {\it Protest Story}}
\vspace{-.2in}
\end{center}
\end{figure*}

\begin{itemize}
\item DramaBank comes with an annotation tool called Scheherezade that produces
the {\sc story intention graph}. Previous research and our own experience suggest
that it is easy to train na\"{\i}ve annotators to annotate stories with Scheherezade. 
\item Scheherezade includes a natural language generator that outputs a retelling
of the original story that reflects the annotators' decisions.
\item Many of the stories share
similar themes, topics and activities that allow us to understand the
common plot structures across stories with similar {\sc sig} representations. 
\item The {\sc sig} allows for the experimentation of producing variations in narrative structure for the same story and exploring stylistic differences in discourse structure and story tellings.
\end{itemize}

We have created the PersonaBank collection of {\sc sig} annotated
stories with several purposes in mind that we believe will make the
corpus useful to other researchers interested in everyday
storytelling, narrative modeling, language generation, and language
processing\footnote{http://nlds.soe.ucsc.edu/personabank}.
Section~\ref{sig-sec} begins by describing the {\sc sig}
representation. Section~\ref{stats} gives a detailed overview of our
corpus. Section~\ref{sch-sec} describes the annotation process used to
generate {\sc sig}s and the challenges of adapting the tool to the
personal narrative domain.  Section~\ref{conc} briefly summarizes
some of the potential applications of the corpus, including
existing work that explores generating many different versions of the
same story given the deep representation of the {\sc sig}. 

\section{Story Intention Graphs}
\label{sig-sec}

{\sc story intention graphs} were built to find computational models that look beyond the surface
form of a text to compare and contrast stories based on content, as opposed to style \cite{elsonthesis}. 
The {\sc story intention graph}, or {\sc sig}, formalism is robust, emphasizing key elements of a narrative rather than attempting to model the entire semantic world of the story. It is an expressive and computable model of content that is accessible for human subject to use an annotation methodology to create an open-domain corpus. 

{\sc sig}s represent a story along several dimensions, starting with
the surface form of the story (first column in
Figure~\ref{squirrel-sig}) and then proceeding to deeper
representations. 
The {\sc sig} bears some similarities to Lehnert's notion of recombinable
plot units as a discourse model
\cite{lehnert1981plot,goyal2010automatically,appling2009representations,nackoul2010text}.
Characters and objects are created, actions and
properties are assigned to them, and interpretations of {\it why}
characters are motivated to take the actions they do are provided. 
The first dimension of the {\sc sig} (second column in Figure~\ref{squirrel-sig})
is called the ``timeline layer'', in which the story facts are encoded
as predicate-argument structures (propositions) and temporally ordered
on a timeline. The timeline layer consists of a network of propositional
structures, where nodes correspond to lexical items that are linked by thematic relations.
The second dimension (third column in Figure~\ref{squirrel-sig}) is
called the ``interpretive layer'' which captures the interpretations of {\it why} characters are motivated 
to take the actions they do. This layer goes beyond
summarizing the actions and events that occur, and attempts to capture
story meaning derived from agent-specific plans, goals, attempts,
outcomes and affectual impacts. Here, the {\sc sig} uses predicates
(discourse relations) that signify plans and goals. 
The final dimension (fourth column in Figure~\ref{squirrel-sig}) is the
``affectual'' layer. Here, affect relations and are represented by the
arcs between story elements. There are a fixed number of types of arcs and affectual nodes that can be used to annotate any kind of story in the interpretation layer, including ``Beowulf'' and ``Gift of the Magi'' \cite{elson2012dramabank}.

For the rest of this paper, we focus on the {\it Startled Squirrel} from Figure~\ref{squirrel-blog-story} and Story 7 from Table~\ref{examples}, which we call the {\it Protest Story}. We provide a brief walkthrough of how to interpret {\sc sig}s ( interpretation arcs are represented in {\bf bold}: 
in the {\it Startled Squirrel} {\sc sig} (Figure~\ref{squirrel-sig}), the narrator places a bowl on the deck (\#1) which {\bf attempts to cause} the goal of the narrator to give the dog some water (\#2) which would {\bf provide for} the dogs' health (a). Then the squirrel approaches the bowl (\#3) to {\bf attempt to cause} the squirrel's goal to drink the water (\#4) which would {\bf provide for} the squirrel's health (b). When the squirrel is started (\#5), this attempts to prevent the goal of drinking the water, and when the squirrel falls (\#6) this both ceases the goal (\#4) and damanges the squirrel's health (b).

In the {\it Protest Story} {\sc sig} (Figure~\ref{protest-sig}), the people are
protesting (\#1) because they disagree about a view the government has
(\#2) (the people disagree which {\bf modifies (causes)} the protest),
thus the peoples' goal is for the government not to rule them, {\bf
  providing for} their health (c) and {\bf damaging} the governments
wealth (d). The protest is {\bf followed by} a rebellion (\#4).

The government has two goals: that the G20 summit succeeds (\#5) and
that the government continue to rule over the people (\#6). A {\bf
  precondition} arc is created to restrict that the goal of the summit
succeeding can only be initialized if the summit has started (\#7). It
is also a {\bf precondition} that the summit must succeed in order for
the government to rule. This would {\bf provide for} the wealth of the
government (a) and the people (b). However, if the government
rules (\#6), this {\bf would prevent} the peoples' goal (\#3).

\section{Overview of the Personal Story Corpus}
\label{stats}

The PersonaBank corpus contains 108 {\sc sig}s created from personal narratives. 
These stories were selected  from the Spinn3r corpus 
and annotated for story topic \cite{icwsm09}.              
We use a lucene index in order to seed topics and rank stories from the 1.5 million stories
\cite{swanson2012say}. We start with 
a list of seeds, for example, for a gardening topic we use [tree, trees, farm, garden, yarn, grass, plant, ...]. 
We get the retrieved list of stories and decide if the story is relevant or not relevant to specified topic.
We assign characteristics to the story that allows us to filter stories we believe are interesting, coherent, overall positive or negative and 
that we believe are possible to encode as {\sc sig}s by considering the following: the narrator is the storyteller, there a clear temporal sequence of events, and the story is not offensive for any reason. 

We select 55 stories that are overall positive and 53 that are negative. 
21 have their interpretation layers annotated. The average number of
words per story is 269 words, and the mimimum and maximum are 104
words and 959 words respectively (Table~\ref{stat-table}). 48 stories have some verbs of communication (e.g. ``said'', ``told''). 
50 stories have an ``in order to'' contingency causal discourse relationship encoded in the {\sc sig}.

\begin{table}[h!]
\begin{center}
\begin{small}
\begin{tabular}{|l|l|}
\hline
Statistics & Stories \\ \hline \hline
Total stories & 108 \\
Positive stories & 55 \\
Negative stories & 53 \\
Interpretation layers annotated & 21 \\
\hline
Average story length in words & 269\\
Minimum & 104 \\ 
Maximum & 959 \\ 
\hline
\end{tabular}
\end{small}
\vspace{-0.1in}
\caption{Overview Statistics of the Personal Story Corpus \label{stat-table}}
\vspace{-0.1in}
\end{center}
\end{table}
\begin{table}[htb!]
\begin{small}
\begin{tabular}{|p{1.0in}|p{1.9in}|}
\hline 
Topic (\#pos, \#neg) & Subtopics (\#pos, \#neg) \\ 
\hline \hline 
Health (1,15) &  Life (1,1),  Death (0,3), Sickness (0,4), Stress (0,2),  Accident (0,3),
Embarrasment (0,2)  \\ \hline
Weather (8,1) & Snow (7,0), Storm (1,1) \\ \hline
Wildlife (10,3) & Squirrels (1,0), Bugs (1,1), Frogs (4,0), Fish (2,0), Birds (0,1), Sharks (1,1), Clams (1,0) \\ \hline
Activities (5,6) &  Photography (1,0), Haircuts (0,4),  Workouts (1,0),  Gardening (2,0), Travel (1,2) \\ \hline
Sports (14,2) &  Swimming (0,1), Scuba (4,0), Fishing (1,0), Running (1,0),  Olympics (1,0),  Camping (3,1), Sledding (4,0) \\  \hline
Holidays and Family (19,4)  & Christmas (7,1), Easter (3,0), Family (9,3)   \\ \hline
Romance (2,22) & New Romance (2,0), Breakups (0,22)  \\ \hline
Everyday Events (10,8) & Dream (1,0), Arrest (0,1) Technology (3,0), Pets (3,1), Work (3,6)  \\ \hline
\end{tabular}
\end{small}
\caption{Topics and Subtopics of Annotated Stories, classified
as to overall affect as being Positive or Negative \label{topics}}

\end{table}
Table~\ref{topics} describes a distribution of the topics that the
stories cover and breaks them down into subtopics. This table also
shows how many of each topic are positive and negative stories. A few
stories cover more than one topic. 
Table~\ref{examples} shows a subset of the corpus stories with their topics, polarity, and an excerpt from the original text. All names have been anonymized with Anne, Jane, Jack and John. 

\begin{table*}[htb]
\centering
\begin{small}
\begin{tabular}{|l|p{.5in}|l|p{4.7in}|}
\hline
Story id & Topic & Polarity & Excerpt \\ \hline \hline
1 & Wildlife, Bugs & POS & Bug out for blood the other night, I left the patio door open just long enough to let in a dozen bugs of various size. I didn't notice them until the middle of the night, when I saw them clinging to the ceiling. Since I'm such a bugaphobe, I grabbed the closest object within reach, and with a rolled-up comic book I smote mine enemies and smeared their greasy bug guts. \\ \hline
7 & Work & NEG & Today was the start of the G20 summit. It happens every year and
it is where 20 of the leaders of the world come together to talk about
how to run their governments effectively and what not. Since there are
so many leaders coming together there are going to be a lot of people
who have different views on how to run the government they follow so
they protest. There was a protest that happened along the street where
I work and at first it looked peaceful until a bunch of people started
rebelling and creating a riot. \\ \hline
17 & Health, Accident, & NEG & So my most recent strangely happy moment. I was in a car accident the other day. Sort of. Heheh. I was heading to Anne's B-day party (mixed with her brothers, they're all born in the same month) and this girl Jane hits me when I'm about to hit the stoplight coming off an exit. I get out of my car, she's spazzing.  \\ \hline
57 & Work, Everyday Events & NEG & Pf changs really messed up my training. It was one person really. If you wouldve seen this schedule i got you would understand. some of you did see it so you know what i mean. I went in last wednesday to take what i thought was my final training class. I was told i missed it and it was the day before. I was really confused becasue my schedule said i was off that day so i pulled it out and showed them. Then she says thats not your schedule. I was like its not what do you mean? \\ \hline
59 & Weather, Snow  & POS & The first day of winter is a huge event, especially for those who are in South Dakota. It is impossible to escape the snow if you're living in SD. For me, this year I was working when it was snowing. I was so sad because I was unable to go out and play in it like I have done ever since I was a child. Fortunately work ended shortly after that and I was able to call up my friends and head to their place.  \\ 
\hline 
\end{tabular}
\caption{Excerpts from PersonaBank \label{examples}}
\vspace{-.1in}
\end{small}
\end{table*}

\section{Scheherazade for Personal Narratives}
\label{sch-sec}

Scheherazade is a freely available annotation tool that facilitates
the creation of {\sc sig}s
\cite{elson2010automatic,elson2012dramabank,elson2012detecting}. 
The annotation process involves sequentially labeling the original story
sentences using a graphical user interface that has been shown to be
usable by na\"{\i}ve annotators. 

We create a new annotation tutorial specifically for personal narratives released with our language resource. We advise annotators that these stories are rich in language and description, and have many things that are not relevant to the annotation of the story events. Before they begin annotating with the tool, they should read the entire story and determine the characters and events that are crucial to the plot. 

The annotation process starts by displaying the original story that is to be annotated (Table~\ref{examples}).
The annotator first defines characters and objects
as props for the story. All the personal narratives are told in a 1st person voice. 
Thus, we define a narrator character to be that voice.
For example, the {\it narrator} and the {\it squirrel} characters in {\it Startled Squirrel} in Figure~\ref{squirrel-blog-story} are defined as ``characters''. The {\it bowl} and {\it deck} were
defined as ``props''. Similarly, the {\it group of leaders}, {\it police} and the {\it protestors} from {\it Protest Story} are defined as ``characters'', and the {\it tear gas} and {\it police cars} are ``props''. 

Next, the annotator assigns actions and properties to the characters and props. 
The annotator highlights segments of text from the original stories and creates story points for each selected
segment by encoding it in predicate-argument structures where nodes correspond to
lexical items that are linked by thematic relations. These story points
create the timeline layer and make up a network of propositional
structures (Column 2 in Figure~\ref{squirrel-sig}). 
\begin{figure}[htb!]
\centering
\includegraphics[width=3.0in]{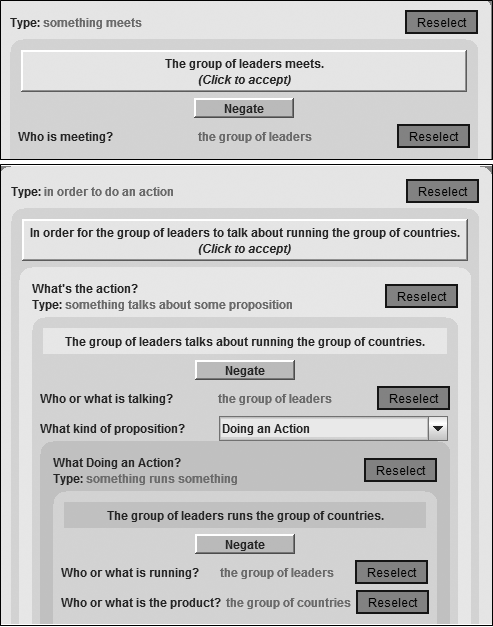}
\caption{\label{sig} Scheherezade annotation screenshot using {\sc wordnet} and {\sc verbnet} lexical resources}
\vspace{-.1in}
\end{figure}

Scheherazade uses the predicate-argument structures from the VerbNet
lexical database \cite{kipper2006extensive} and uses WordNet
\cite{fellbaum1998} as its noun and adjectives taxonomy.
Figure~\ref{sig} shows a screenshot of the Scheherezade annotation
tool, illustrating the process of assigning propositional
structure to the sentence ``20 of the leaders of the world come together to talk about how to run their government effectively and what not" from {\it Protest Story}. 
We want to annotate the idea of ``coming together" or ``meeting", and decide on the idea of ``a group of leaders coming together to discuss how to run their governments''. 
We encode this in Scheherazade as two propositions: {\tt meet(group of leaders)} and {\tt in order to(talk about running(the group of countries))}. We begin by typing ``meets" into the Scheherazade GUI (Figure~\ref{sig}).  A list of possible verb senses and arguments from VerbNet are shown. The annotator selects the one most appropriate for their text span. After selecting a sense, the GUI displays the slots within the frame that need to be filled. In this example, the GUI asks `who is meeting' and we select ``the group of leaders" from our characters we previously defined. Next, we encode a nested proposition as the prepositional phrase {\tt in order to(talk about running(the group of
  countries))}. Both actions ({\tt meet} and {\tt in order to talk})
contain references to the story characters and objects ({\tt group of
  leaders} and {\tt group of countries}) that fill in slots
corresponding to semantic roles.

The annotation process is facilitated by the fact that
Scheherazade includes a built-in language generation module that
helps na\"{\i}ve users produce correct annotations by automatically
generating the natural language realization of their encoding
incrementally as they annotate the story
\cite{bouayad1998integrating}. 
This is called a what-you-see-is-what-you-mean ({\sc wysiwym}) paradigm.
The result of our annotation for the propositions  {\tt meet(group of leaders)} and {\tt in order to(talk about running(the group of countries))} is {\it The group of leader meet in order to
  talk about running the group of countries}. 
Figure~\ref{protest-sch} and~\ref{squirrel-sch} show the entire realization for
the {\it Protest Story} and {\it Startled Squirrel}. We also include the
Scheherezade outputs for each story in the corpus.

WordNet and VerbNet provide a great deal of flexibility 
in the underlying semantic representation of the {\sc sig} timeline.
For example, it is easy to define
and name objects like ``G20 summit'' for the meeting in {\it Protest Story}. 
Because the {\sc sig} representation uses a
discourse model of the characters and props in the story, the
{\sc wysiwym} realizer also automatically understands entities for
coreference. For example, the {\it Startled Squirrel} describes a group of 
squirrels and then an individual crazy squirrel. The realizer knows the 
difference between both squirrels, realizing the first as ``a group of squirrels'' 
and the individual squirrel as ``a crazy second squirrel''  so
they can be refered to as different entities.

\begin{figure}[htb]
\centering
\begin{small}
\begin{tabular}{|p{2.9in}|}
\hline
There once was a group of leaders.  The group of leaders was meeting in order to talk about running a group of countries and near a workplace. A group of peoples protested because the group of leaders was meeting and began to be peaceful. A group of peoples stopped being peaceful, began to be riotous, burned a group of police cars and a group of peoples pelted a group of police officers. A group of police officers wore some armor because a group of peoples was riotous. A group of peoples smashed a group of windows of a group of stores, and a group of police officers attacked a group of peoples with a tear gas and with a group of riot guns. \\
\hline
\end{tabular}
\caption{Scheherazade Realization of {\it Protest Story} \label{protest-sch}}
\vspace{.1in}
\begin{tabular}{|p{2.9in}|}
\hline
A narrator placed a steely and large bowl on a back deck in order for a dog to drink the water of the bowl. The bowl began to be popular, and a group of birds drank the water of the bowl and bathed the group of birds in the bowl. The group of birds organized itself on the railing of the deck and in order to wait. A group of squirrels drank the water of the bowl. A crazy second squirrel approached the bowl. The second squirrel began to be startled because it saw the reflection of the second squirrel. The second squirrel leaped because it was startled and fell over the railing of the deck and because it leaped. The second squirrel held the railing of the deck with a paw of the second squirrel. The paw of the second squirrel slipped off the railing of the deck, and the second squirrel fell. \\
\hline
\end{tabular}
\caption{Scheherazade Realization of {\it Startled Squirrel} \label{squirrel-sch}}
\vspace{-.1in}
\end{small}
\end{figure}

When annotating personal narratives, we find difficulties in adapting
the annotation tool to this new domain. Personal narratives might
contain some descriptive parts that are not easy to annotate and
interpret using {\sc sig} representation. These descriptions mostly do
not pertain to the key aspects of the story.
For example, Story 57 begins with the following sentences:
``Pf changs really messed up my training. It was one person really. If you wouldve seen this schedule i got you would understand. some of you did see it so you know what i mean.'' 
These are observations that aren't critical to
the events of the story. The action starts at the fifth sentence: 
``I went in last wednesday to take what i thought was my final training class.'' We encourage
annotators to ignore descriptive observations in texts such as these
if they are not central to the action of the narrative.

There are other situations where annotators cannot find the
exact words or expressions from the original story in the WordNet or VerbNet dictionaries. 
We encourage them to choose an appropriate paraphrase that conveys
the same concept. For example, to annotate the phrase ``There was a protest that
happened'', note there are many possible propositional representations
of this event. Our annotator selected the proposition``the people protested because the group of
leaders was meeting''. As another example, the expletive ``it" as in
``It was hard to...'' can be represented instead as ``The situation
was hard to...''. The blogs often discuss events the narrator was
involved with, and use the pronoun ``we''. This cannot be
annotated in WordNet or VerbNet, so to represent ``we decide to...'' we can annotate it as
``A group of friends decided to...'' where we choose an appropriate
group of characters based on the context of the story. Similarly, we
can use groups for plurals. Instead of ``five trees'' we represent ``a
group of trees''.

There are many possible interpretations of these stories, and thus many possible annotations. For the phrase ``There was a protest that happened'' we may instead decide to annotate it as ``the people protested against the group of leaders because the group of leaders was meeting'' or ``the people protested against the group of leaders because the people disagree about an ideality''. We may instead choose a different verb for the deep representation such as ``the people disagree about an ideality'', ``the people dislike the government'', or ``the people distrust the government because the people disagree about an ideality''.

All our stories were annotated once by expert annotatators. We did not believe annotator agreement was an important task for us because there are so many ways to interpret a story. The DramaBank contains multiple encodings for
some of the Aesop's Fables. \cite{elson2012dramabank} examines annotator agreement for Aesop's Fables and found this a very difficult task to measure. We imagine it would be even more difficult with the complexity of personal narratives. Instead, we ensure that each annotation is very rich and complete in itself. 
The final realization reflects one annotators' interpretation of the story. 

Our expert annotators can annotate the timeline layer of a story in about one hour. Annotating the interpretive and
affectual layers requires more subjective judgement and takes an additional hour for each story. 
We review the annotations with our annotators until they feel comfortable with
the annotation process. We find that annotation is facilitated by
using annotators that have a background in linguistics and most of our
annotations have been produced by linguistics undergraduates working
as research assistants. We are continually adding new stories
to the corpus.

\section{Applications}
\label{conc}

To date, there are several projects that
have made use of the {\sc sig} representation for applications related
to storytelling, game playing and narrative generation.
\cite{harmon2015imaginative} draw parallels between the narrative
representation of {\sc sig}s and Skald, a narrative generator. This
combination allows for narrative generation while keeping the
affordances of the {\sc sig} representation. 
{\sc sig}s are also hugely beneficial as a content planner. 
\cite{antoun2015generating} has used the {\sc sig} as
an intermediate representation of meaning by transforming a play
trace of the PromWeek game into a representation which can be used to
generate natural language recaps of the game. 

\begin{table}[ht!]
\centering
\begin{small}
\begin{tabular}{|p{3.0in}|}
\hline 
\begingroup
Original excerpt: I went in last wednesday to take what i thought was my final training class.
\endgroup
\\ \hline
\begingroup
{\sc wysiwmy} excerpt: A resolute narrator named Anne excitedly entered a restaurant in order for a disgruntled manager to train the narrator.  
\endgroup
\\ \hline
\begingroup
{\sc est} output: I excitedly entered PF Changs in order for the manager to train me.
\endgroup
\\ \hline
Syntactic structure:
\begingroup
\fontsize{7pt}{7pt}
\begin{alltt}
<dsynts id="0">
<dsyntnode class="verb" lexeme="enter" 
  mode="" mood="ind" rel="II" tense="past" 
  wn_offset="2016523">
    <dsyntnode article="no-art" class="common_noun" 
    gender="fem" lexeme="narrator" number="sg" 
    person="1st" pro="pro" rel="I"/>
    <dsyntnode article="no-art" class="proper_noun" 
    gender="neut" lexeme="PF Changs" number="sg" 
    person="" rel="II" wn_offset="4081281"/>
    <dsyntnode class="preposition" 
    lexeme="in_order" rel="ATTR">
      <dsyntnode class="verb" extrapo="+" 
      lexeme="train" mode="inf-to" mood="inf-to" 
      rel="II" tense="inf-to" wn_offset="603298">
        <dsyntnode article="def" class="common_noun" 
        gender="fem" lexeme="manager" number="sg" 
        person="" rel="I" wn_offset="10014939"/>
        <dsyntnode article="no-art"
        class="common_noun" gender="fem" 
        lexeme="narrator" number="sg" 
        person="1st" pro="pro" rel="II"/>
      </dsyntnode>
    </dsyntnode>
    <dsyntnode class="adverb" lexeme="excitedly" 
    position="pre-verbal" rel="ATTR"/>
  </dsyntnode>
</dsynts>
\end{alltt} 
\endgroup
\\ \hline
\end{tabular}
\caption{Excerpt from Story 57, and corresponding {\sc wysiwym} realization, output from {\sc est} and deep syntactic structure. \label{dsynts}}
\end{small}
\vspace{.1in}
\centering
\begin{small}
\begin{tabular}{|l|p{2.6in}|}

\hline
Num & Sentence Variations \\ \hline \hline
1 & I rather excitedly entered PF Changs because the manager wanted to train me.  \\ \hline
2 & The manager wanted to train me, so I excitedly entered PF Changs, okay? \\ \hline
3 & I excitedly entered PF Changs in order for the manager to train me. \\ \hline
4 & The manager wanted to train me, so I excitedly entered PF Changs.	 \\ \hline
5 & Ok, I excitedly entered PF Changs in order for the manager to train me, right?  \\ \hline
6 & Because the manager wanted to train me, I excitedly entered PF Changs. \\ \hline
7 & The manager wanted to train Anne, so she excitedly entered PF Changs, as it were. \\ \hline
8 & Because the manager wanted to train Anne, she excitedly entered PF Changs!!  \\ \hline
9 & Anne excitedly entered PF Changs  \\ \hline
10 & Essentially, ok, the manager wanted to train Anne, so she excitedly entered PF Changs. \\ \hline
11 & Actually, Anne excitedly entered PF Changs in order for the manager to train her. \\ \hline
12 & The director wanted to train Anne, so she excitedly entered PF Changs.	 \\ \hline 
\end{tabular}
\caption{Sentence Variations from Story 57 \label{sent1}}
\end{small}
\vspace{-.2in}
\end{table}

Our companion paper \cite{huetal16-lrec} released the Story Dialogue
with Gestures (SDG) corpus which contains 50 personal narratives
rendered as dialogues between two agents with complete gesture and
placement annotations. Their first approach for generating dialogues
manually split personal narratives. Their second approach uses
{\sc sig}s of personal narratives as a way to get story content
from the {\sc wysiwym} realization and automatically produce dialogues
from this telling, using the {\sc est}.

The Expressive-Story Translator ({\sc est}) explores the use of
personal narratives in storytelling by utilizing the rich
representation of {\sc sig}s. Storytellers dynamically adjust their
narratives to the context and their audience, telling and retelling
the same story in different ways depending on the listener and the
particular communicative goal or intention and ``explore'' incidents
by offering many interpretations of the same incident
\cite{mateas04preliminary}. For example, storytellers tell richer
stories to highly interactive and responsive addressees
\cite{thorne1987study}, and stories told by young adults ``play'' to
the audience, repeatedly telling a story to get a desired effect and
communicate effectively with the audience \cite{thorne2003telling}.

\cite{rishes2013generating} use {\sc sig}s of Aesop's Fables from the
DramaBank for retelling these stories in different ways by creating a
mapping between the content representation of the {\sc sig} and the
syntactic representation used by the {\sc personage} natural language
generation engine \cite{mairesse2007personage}. This syntactic
representation of a story enables the retelling of any story that is
represented as a {\sc sig}. Table~\ref{dsynts} shows an excerpt from
Story 57, the corresponding {\sc wysiwmy} realization, and the
generated output from the {\sc est} and the syntactic structure that
allows the {\sc est} to generate variations.

PersonaBank allows for the exploration of
storytelling by retelling personal narratives by implementing
a variety of narrative and sentence parameters including
contingency discourse relations into the generation of
stories
\cite{lukin2015narrative,lukin2015generating}. Our current work
examines many possible combinations of parameters and generates many
versions of a sentence according to different framing goals.  For
example, Table~\ref{sent1} shows
how the {\sc est} generates surface strings from the syntactic template from
Table~\ref{dsynts} and how we can create many variations by utilizing
generation parameters. Variations 1 thru 6 are in the first person perspective,
and 7 thru 12 are in the third person voice and reference the narrator as
Anne. All the sentences vary the sentence construction in some
way. Sentences 1, 2, 5, 7, 8, 10, 11, and 12 include additional
variations provided by Personage parameters, including hedging,
acknowledgements, synonyms, and exclamation marks.

The {\sc sig} offers rich information in the interpretation layer
that has not yet been utilized in existing work. We plan to derive
character emotions from this layer according to appraisal theory. With
a rich understanding of the impact of actions of characters in the
story, this information can be used by a variety of expressive methods
to enhance the storytelling experience including using more expressive
language, gestures, and speech influenced by emotions.

Other potential applications of the {\sc sig} and storytelling include
question and answering from {\sc sig}s about the story domain. Because
the {\sc sig} provides all the world knowledge about the story, it is
possible to ask questions about the narrative structure, events that
happened, and the effects they had on characters and their plans.

\section{Conclusion}
We have described the PersonaBank corpus, a corpus of
personal narratives that have been annotated with the {\sc story
  intention graph} representation for stories as used in
the DramaBank language resource.
We believe this corpus will be of general utility both for theoretical
analyses of narrative structure and for applications related to
storytelling and dialogue.

\section{Acknowledgements}
This research was supported
by Nuance Foundation Grant SC-14-74,
NSF Grants IIS-HCC-1115742 and IIS-1002921.

\section*{Appendix: Story Intention Graphs}
\begin{figure*}[h!]
\begin{center}
\centering
\includegraphics[width=5in]{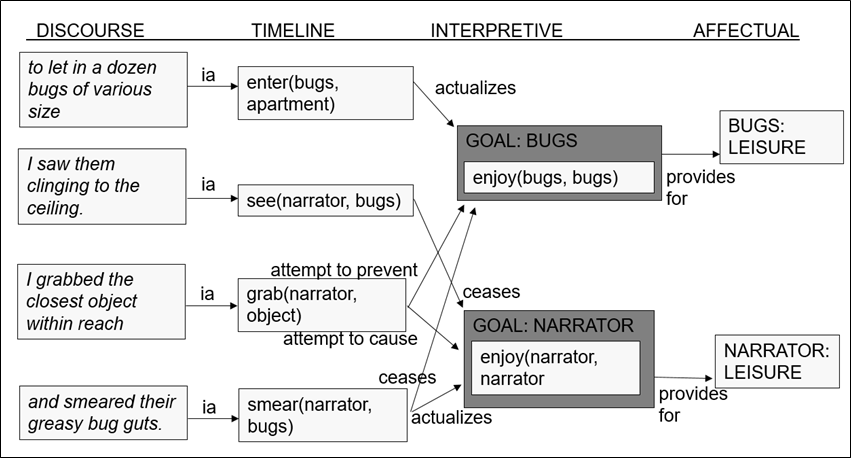}
\vspace{-.1in}
\caption{\label{bug-sig} Part of the {\sc Story Intention Graph} for Story 1}
\vspace{.2in}
\includegraphics[width=5in]{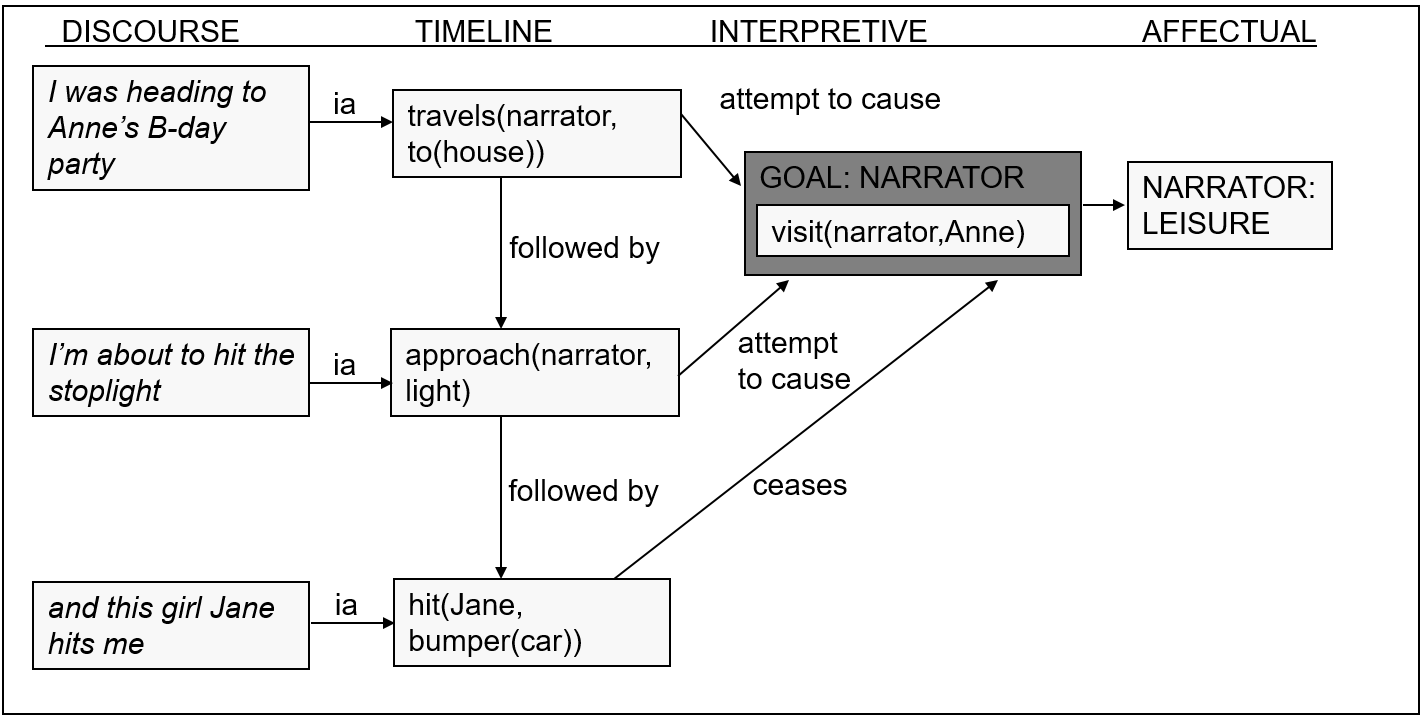}
\vspace{-.1in}
\caption{\label{17-sig} Part of the {\sc Story Intention Graph} for Story 17}
\vspace{.2in}
\includegraphics[width=5in]{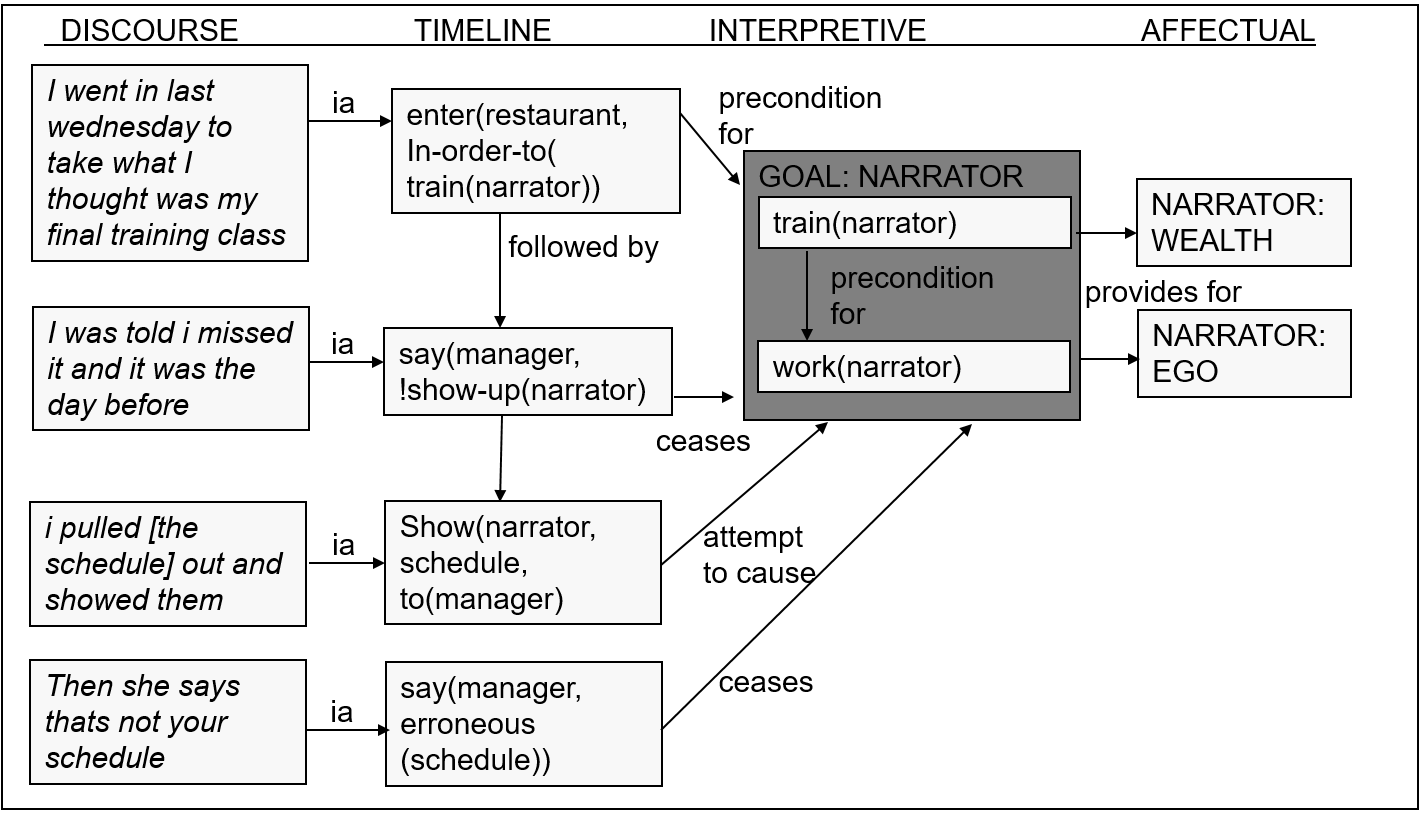}
\vspace{-.1in}
\caption{\label{57-sig} Part of the {\sc Story Intention Graph} for Story 57}
\end{center}
\end{figure*}

The Appendix provides three more {\sc sig}s corresponding to Story 1 (Figure~\ref{bug-sig}), Story 17 (Figure~\ref{17-sig}), and Story 57 (Figure~\ref{57-sig}) from Table~\ref{examples}, illustrating the generality of the {\sc sig} modeling to be applied to any domain. 

\nocite{*}
\section{Bibliographical References}
\label{main:ref}
\bibliographystyle{lrec2016}
\bibliography{lrec2016lukin}

\end{document}